%% file: main.tex
\documentclass{article}

 \usepackage[preprint,nonatbib]{neurips_2026}

\usepackage[utf8]{inputenc} 
\usepackage[T1]{fontenc}    
\usepackage{hyperref}       
\usepackage{url}            
\usepackage{booktabs}       
\usepackage{amsfonts}       
\usepackage{nicefrac}       
\usepackage{microtype}      
\usepackage{xcolor}         

\usepackage[style=numeric-comp, maxbibnames=99, minbibnames=99]{biblatex} 
\usepackage{amsmath}

\usepackage{nicematrix}

\usepackage{enumitem}
\setlist{itemsep=0pt, topsep=0pt}

\usepackage{siunitx}

\usepackage{wrapfig}

\usepackage{microtype}

\renewcommand{\paragraph}[1]{\vspace{.5em}\noindent\textbf{#1}}

\usepackage{xspace}
\makeatletter
\DeclareRobustCommand\onedot{\futurelet\@let@token\@onedot}
\def\@onedot{\ifx\@let@token.\else.\null\fi\xspace}
\def\eg{\emph{e.g}\onedot} 
\def\ie{\emph{i.e}\onedot}

\makeatother

\title{SapiensID 2.0: Aligning Human Recognition Foundation Models with Human Perception}

\author{%
    Yiyang Su\textsuperscript{1}\quad
    Jie Zhu\textsuperscript{1}\quad
    Feng Liu\textsuperscript{2}\quad
    Anil Jain\textsuperscript{1}\quad
    Xiaoming Liu\textsuperscript{1,3}\\
    \textsuperscript{1} Michigan State University\quad
    \textsuperscript{2} Drexel University\quad
    \textsuperscript{3} University of North Carolina, Chapel Hill\\
    {\tt\small \{suyiyan1, zhujie4, jain\}@msu.edu, fl397@drexel.edu, liuxm@cs.unc.edu}
}

\begin{document}

\maketitle

\input{secs/0_abstract}
\input{secs/1_intro}
\input{secs/2_related_work}
\input{secs/3_method}
\input{secs/4_experiments}
\input{secs/5_conclusion}

{
\small
\printbibliography
}


\newpage
\appendix

\input{secs/X_supp}



\end{document}

%% file: secs/0_abstract.tex
\begin{abstract}
While foundation models have significantly advanced human recognition across diverse modalities, they predominantly rely on static, geometric feature extraction.
This approach fundamentally diverges from human perception.
Consequently, current models often suffer from ``semantic blindness,'' overfitting to transient noise while failing to leverage invariant soft biometrics, and struggle to capture temporal motion signatures.
To bridge this gap, we propose SapiensID 2.0, a human recognition framework enriched with both semantic and temporal awareness.
To overcome the lack of soft-biometric annotations, we transfer zero-shot semantic knowledge from Multimodal Large Language Models (MLLMs) into a discriminative embedding space.
We resolve the dimensional mismatch between these spaces using Invariant Trait Alignment (ITA) to distill core persistent traits, and Transient Noise Disentanglement (TND) to decouple artifacts like clothing.
Furthermore, we design a Kinematic Semantic Attention Head (K-SAH) that extends spatial attention across temporal windows.
By tracking semantic patches over time, K-SAH captures rich kinematic signatures without requiring large-scale video datasets.
Extensive experiments demonstrate that SapiensID 2.0 achieves state-of-the-art performance across image- and video-based person re-identification and gait recognition, while maintaining robust face recognition capabilities.
\end{abstract}

%% file: secs/1_intro.tex
\section{Introduction}
Human recognition systems~\cite{minaee2023biometrics} typically isolate different modalities into distinct, modality-specific models.
SapiensID~\cite{kim2025sapiensid} marked a paradigm shift by using a single model for face recognition and person re-identification (reID).
In real-world scenarios, SapiensID can dynamically fall back on the other modality when extreme poses, poor visibility, or occlusions compromise a single modality (\eg, an obscured face).
However, SapiensID extracts discriminative biometric features from raw visual textures of individual frames, and its embedding space inherently lacks explicit semantic awareness and kinematic continuity.
In contrast, human recognition is anchored precisely by these two pillars.

\begin{figure}
    \centering
    \includegraphics[width=\linewidth]{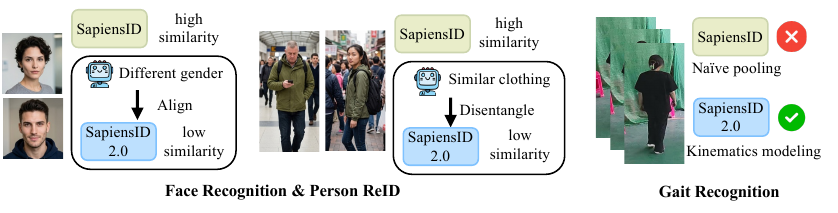}
    \vspace{-2em}
    \caption{\small SapiensID~\cite{kim2025sapiensid}, based on face recognition and person reID, mistakenly assigns high similarity to pairs, since it does not consider fixed soft biometrics like race and heavily relies on dynamic soft biometrics like clothing.
    Conversely, our approach, SapiensID 2.0, utilizes information extracted by MLLM from images to align fixed soft biometrics with biometric features and disentangle noise to achieve more robust human recognition.
    Additionally, we extend SapiensID to gait recognition by explicitly modeling the subject's kinematics in videos.
    \vspace{-5mm}}
    \label{fig:teaser}
\end{figure}

First, humans rely on interpretable semantic attributes~\cite{hassan2024soft}.
We anchor identity to \textit{fixed soft biometrics}---anatomically invariant traits such as race, gender, and body proportions.
Also, we actively discount \textit{dynamic soft biometrics}---transient, variable traits such as clothing or accessories.
As shown in Fig.~\ref{fig:teaser}, standard metric learning models exhibit a critical ``semantic blindness.''
SapiensID can confidently assign high similarity to two face images of different races or to two whole-body images because they share similar clothing.
The model relies on entangled visual textures rather than distinguishing between explicit, identity-preserving attributes and transient noise.

Second, human perception naturally leverages \textit{kinematic continuity}---the consistent, temporal evolution of a subject's structural mechanics (\eg, gait) while in motion.
SapiensID, however, operates entirely on static frames.
When applied to video-based tasks such as gait recognition, it extracts features for each frame and averages the features across frames.
This static formulation completely discards the rich, discriminative kinematic signatures inherent in human movement.

Elevating foundation models to capture semantic awareness and kinematic continuity presents a data scarcity challenge.
On the semantic front, large-scale reID datasets like WebBody~\cite{kim2025sapiensid} lack soft biometric annotations such as race, gender, and clothing.
They are collected via keyword queries and face similarities, which do not provide semantic labels.
On the temporal front, WebBody consists of more than 90\% static images.
Compiling and labeling a video dataset of comparable scale to learn kinematic signatures from scratch is not feasible.

To overcome the semantic annotation bottleneck, we leverage multimodal large language models (MLLMs, \eg, \cite{bai2025qwen2}), which can generate accurate pseudo-ground truths for both fixed and dynamic soft biometrics.
However, deploying generative MLLMs directly at inference is computationally infeasible.
We must therefore transfer their deep semantic knowledge to construct an enhanced, more discriminative feature space.
This transfer introduces a structural mismatch: semantic attributes are inherently low-dimensional signals (\eg, age, a scalar, or race, categorical), whereas MLLM embeddings are highly entangled and high-dimensional.
Directly aligning these spaces would force the biometric model to absorb irrelevant linguistic artifacts.

To resolve this, we propose a Bipartite Subspace Alignment strategy.
For fixed soft biometrics, we introduce Invariant Trait Alignment (ITA), which distills the core variance of stable identity traits into the biometric space.
Conversely, to handle dynamic soft biometrics, we introduce Transient Noise Disentanglement (TND), which enforces strict orthogonality against the principal components of transient traits, effectively neutralizing clothing bias.

Simultaneously, to model kinematic continuity without requiring a massive, newly labeled video dataset, we introduce the Kinematic Semantic Attention Head (K-SAH).
Rather than training video architectures from scratch, K-SAH harnesses the temporal modeling potential of our image-trained model.
By extending the existing semantic attention mechanism across a temporal window, the network explicitly tracks the spatial evolution of the same semantic patches (\eg, body joints) across sequential frames.
Crucially, K-SAH is versatile: it exploits kinematic motion patterns when presented with video tracklets, while degenerating to spatial-only processing for static images.

By successfully integrating semantic awareness and kinematic continuity, our proposed architecture, SapiensID 2.0, serves as a unified human recognition foundation model across face recognition, person reID, and gait recognition.
Extensive experiments demonstrate that SapiensID 2.0 significantly outperforms the previous state of the art from video- to image-based person reID, from short- to long-term reID, and maintains robust face recognition capabilities.

Our contributions are as follows:
\begin{itemize}[leftmargin=2em]
\item \textbf{Semantic Subspace Alignment via Distillation}: We propose a feature-based knowledge distillation framework for semantic subspace alignment using MLLMs.
It employs mathematically distinct pathways---ITA and TND---to simultaneously distill fixed soft biometrics and purge dynamic soft biometric noise from the discriminative feature space.
\item \textbf{Kinematic Semantic Attention Head (K-SAH)}: We adapt spatial semantic attention to process temporal sequences via tracked structural patches and window attention.
This architecture elegantly captures kinematic continuity when video is available without extensive video training, while degenerating to standard static processing for individual images.
\item \textbf{A Unified Model for Human Recognition}: By integrating semantic awareness and kinematic continuity, SapiensID 2.0 provides a unified foundation model spanning face, reID, and gait recognition, achieving superior performance in generalized recognition tasks.
\end{itemize}

%% file: secs/2_related_work.tex
\section{Related Work}

\paragraph{Human Recognition.}
Conventional biometric systems isolate human recognition into specialized modalities, \eg, face recognition~\cite{ebrahimi2025gif, you2025lvface, peng2025stylized, kim2025vigface, kim2025idface} and person reID~\cite{su2025hamobe,ha2025multi,son2025towards,li2025one,xu2025self,zuo2025reid5o,pang2025identity,yuan2025poses,wang2025secap, cui2025dkc, liang2025differ, chen2025learning, khalid2025bridging, zhang2025viperson}.
While multi-task frameworks attempt to fuse these domains~\cite{liu2024farsight,liu2025person,zhu2025quality}, or explore joint video-image pretraining (\eg, VILLS~\cite{huang2025vills}), they either rely on late-stage decision-level score fusion~\cite{zhu2025quality} or require joint video training while omitting face recognition~\cite{huang2025vills}.
SapiensID~\cite{kim2025sapiensid} established a unified representation paradigm by training on massive body-centric data, yielding robust, pose-invariant features across face and body.
Despite this architectural unification, SapiensID remains semantically agnostic and temporally static---frequently making erroneous geometric matches that contradict obvious human attributes.
SapiensID 2.0 evolves this foundation through: (1) grounding the embedding space in explicit semantics distilled from MLLMs, and (2) extending spatial attention into the temporal domain to dynamically track tokens across frames.

\paragraph{Soft Biometrics.}
Soft biometrics have proven effective~\cite{abdelwhab2018survey} in improving face~\cite{jain2009facial,park2010face,tome2015facial}, and fingerprint recognition~\cite{ailisto2006soft,jain2004integrating} accuracy.
However, their reliance on manually annotated attributes severely limits their experimental scale.
In clothing-changing person reID~\cite{yang2019person,gu2022clothes}, disentangling identity from transient appearance relies on clothing labels.
Recently, DIFFER~\cite{liang2025differ} addressed this using an adversarial Gradient Reversal Layer, which can be unstable during training.
For foundation models, training stability is even more important.
In contrast, our framework avoids adversarial training and transfers zero-shot MLLM knowledge purely via a stable, feature-based distillation.
By employing ITA to directly embed fixed anatomical traits and TND to explicitly orthogonalize dynamic clothing noise, our approach achieves robust semantic disentanglement.

\paragraph{Gait Recognition.}
Gait recognition identifies individuals based on their walking posture, using modalities such as appearance~\cite{habib2025cargait,wang2025gait,huang2025learning,huang2026vocabularyguided,jin2025denoising,yang2025bridging}, skeletons~\cite{fan2024skeletongait}, and LiDAR~\cite{shen2025lidargait++}.
The majority of methods---whether utilizing CNNs~\cite{fan2023opengait}, transformers~\cite{fan2023exploring}, or large vision models~\cite{ye2024biggait,ye2025biggergait}---require large-scale video datasets for training and inference.
A separate line of research has explored image-based or static-to-dynamic approaches.
Early methods compress entire gait cycles into static templates, such as Gait Energy Images (GEIs), though these suffer from severe information loss.
\cite{xu2020gait} attempts to reconstruct a full gait sequence from a single image during inference; however, training their reconstruction module still requires video data.
While general spatio-temporal video transformers (\eg, ViViT~\cite{arnab2021vivit}) rely on dense grid-based self-attention with quadratic complexity and require video fine-tuning, we demonstrate that a foundation model trained primarily on images can inherently capture temporal dynamics.
By adapting the spatial attention mechanism to track semantically anchored keypoint tokens across neighboring frames, our approach captures rich kinematic signatures directly without video retraining.

%% file: secs/3_method.tex
\section{Method}

\subsection{Preliminaries}


\begin{figure}
    \centering
    \includegraphics[trim={0in 0in 0in 0.34cm},clip,width=4.89in]{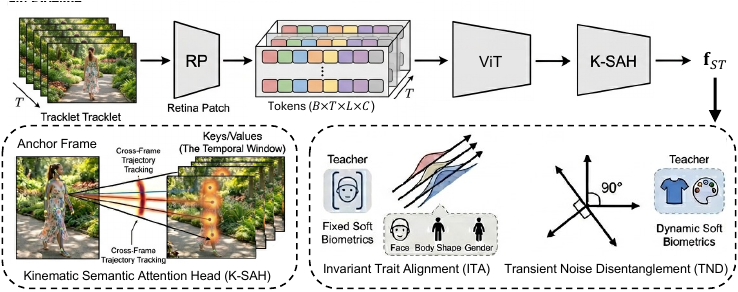}
    \vspace{-0.6em}
    \caption{\small Overall architecture of SapiensID 2.0.
    RetinaPatch (RP) converts the image into patches based on the landmarks and key points.
    ViT extracts features from these patches.
    K-SAH aggregates the features around each joint in images or videos.
    Additionally, we supervise the training via alignment with fixed soft biometrics and the disentanglement of dynamic soft biometrics.
    \vspace{-3mm}}
    \label{fig:pipeline}
\end{figure}

SapiensID~\cite{kim2025sapiensid} uses Retina Patch (RP) to dynamically allocate patches to semantically critical regions (\ie, face, upper torso, whole body).
RP generates semantically grounded patches with region-sampled positional embeddings that preserve spatial origin---an inductive bias critical for temporal tracking.
These patches pass through the ViT~\cite{ViT} backbone to extract a deep spatial feature map $\mathbf{X} \in \mathbb{R}^{L \times C}$, where $L$ is the number of spatial tokens and $C$ is the feature dimension.

Following the ViT, the Semantic Attention Head (SAH) aggregates this feature map.
SAH generates a semantic query matrix $\mathbf{Q} \in \mathbb{R}^{K \times C}$ from predicted keypoints using a 2D position embedding matrix $\mathbf{P} \in \mathbb{R}^{L \times C}$:
\begin{align}
    \mathbf{Q} = \text{GridSample}(\mathbf{P}, \mathbf{C}) + \mathbf{O},
    \label{eq:image_grid_sample}
\end{align}
where $K$ is the total number of semantic queries ($K=17$ keypoints), $\mathbf{C} \in \mathbb{R}^{K \times 2}$ represents the image-specific predicted 2D coordinate matrix, and $\mathbf{O} \in \mathbb{R}^{K \times C}$ is a learned global offset matrix.
These queries extract local features via a cross-attention operation over the spatial tokens:
\begin{align}
    \mathbf{H} = \text{Attention}(\mathbf{Q}, \mathbf{P}, \mathbf{X}).
    \label{eq:image_semantic_attention}
\end{align}
Here, \(\mathbf{P}\) serves as the key and \(\mathbf{X}\) serves as the value.
This allows the network to adaptively pool features around keypoints.
$\mathbf{H}$ is then flattened to produce the final biometric feature $\mathbf{f}$.

However, SapiensID has two limitations.
First, its feature space lacks explicit awareness of fixed soft biometrics and is vulnerable to dynamic soft biometrics.
Second, when given a video, it extracts biometric features for each frame and collapses them via averaging.
This isolates the spatial attention (Eq.~\ref{eq:image_semantic_attention}) to individual frames and discards the kinematic continuity inherent in temporal dynamics.

\subsection{Overall Framework Formulation}

To align the latent space with human cognitive identification, we propose SapiensID 2.0 (see Fig.~\ref{fig:pipeline}), a unified architecture driven by explicit soft biometrics and temporal dynamics.
Like SapiensID, given a video, we continue to use RP and a ViT backbone to extract structurally grounded features \(\mathbf{X}\).

Our objective is to extract an enhanced spatio-temporal feature $\mathbf{f}_{ST}$ that satisfies two novel constraints:

\begin{enumerate}[leftmargin=2em]
    \item \textbf{Bipartite Semantic Alignment}: The feature $\mathbf{f}_{ST}$ must explicitly capture the variance of \textit{fixed soft biometrics} while decoupling from \textit{dynamic soft biometrics}.
    Its projection must be highly correlated with the underlying semantic subspace of fixed soft biometrics and strictly orthogonal to the confounding subspace of dynamic soft biometrics.
    \item \textbf{Kinematic Continuity}: The network must explicitly model the trajectory of the semantic queries $\mathbf{Q}$ across the temporal dimension $T$, capturing the continuous kinematic signatures (\eg, gait) that humans use for identification.
\end{enumerate}

We achieve the first constraint through a novel feature-based distillation module incorporating ITA and TND.
To address the second constraint, we replace the static SAH with our K-SAH, which extends cross attention pooling over temporal windows.

\subsection{Semantic Alignment via Distillation}

\begin{figure}
    \centering
    \includegraphics[width=\linewidth]{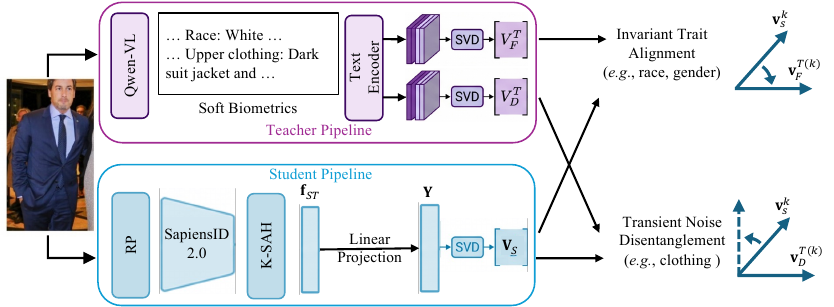}
    \vspace{-1.5em}
    \caption{\small Semantic alignment via distillation.
    ITA aligns the principal components of the \(\mathbf{Y}\) with the principal components of fixed soft biometric embeddings in the MLLM space.
    TND enforces that the principal components of the \(\mathbf{Y}\) are orthogonal to those of the confounding factors like clothing, thereby minimizing their influence.
    \vspace{-3mm}}
    \label{fig:soft_biometrics}
\end{figure}

Deploying a generative MLLM at the edge for real-world applications incurs prohibitive computational overhead.
Thus, we propose a feature-based cross-modal distillation framework to efficiently transfer high-dimensional semantic understanding into the SapiensID student model (see Fig.~\ref{fig:soft_biometrics}).

\paragraph{Semantic Subspace Construction.}
To generate the requisite semantic supervision, we query an MLLM offline on the WebBody4M dataset.
It outputs attributes in two groups: Fixed Soft Biometrics (\eg, race, gender) and Dynamic Soft Biometrics (\eg, clothing, hair).
These textual descriptions pass through the frozen text encoder, yielding high-dimensional semantic embeddings representing the native hidden spaces: $\mathbf{X}^{F}$ for fixed soft biometrics and $\mathbf{X}^{D}$ for dynamic soft biometrics.

To bridge the dimensional gap between the visual metric space and the language space during training, the unified spatio-temporal biometric feature $\mathbf{f}_{ST}$ is passed through a shallow Multi-Layer Perceptron (MLP).
This yields a unified projected student feature $\mathbf{Y}$.
We mathematically force this feature to simultaneously align with invariant traits and orthogonalize against transient noise.
This MLP and the teacher text encoder are utilized only for loss calculation during training and are completely discarded during inference.
During deployment, SapiensID 2.0 operates strictly on the 63M parameter ViT-B backbone in a single forward pass, introducing zero additional computational overhead.

\paragraph{Invariant Trait Alignment (ITA).}
Fixed soft biometrics are low-dimensional signals representing stable identity markers.
To embed this knowledge, ITA aligns the orthogonal basis vectors containing the highest semantic variance of the teacher's fixed subspace with the SapiensID feature space.
By calculating the empirical covariance matrix of $\mathbf{X}^{F}$ and performing Singular Value Decomposition (SVD), we extract the top $D_F$ eigenvectors, represented by the matrix $\mathbf{V}_T^F \in \mathbb{R}^{C \times D_F}$.
A parallel decomposition on the projected student feature $\mathbf{Y}$ yields its principal components matrix, $\mathbf{V}_S$.
The alignment loss is:
\begin{align}
    \mathcal{L}_{ITA} = \sum_{k=1}^{D_F} \left( 1 - \left\langle \mathbf{v}_{T}^{F(k)}, \mathbf{v}_{S}^{(k)} \right\rangle^2 \right) + \lambda_{P} \left\lVert \mathbf{X}^F \mathbf{V}_T^F - \mathbf{Y} \mathbf{V}_S \right\rVert_F^2,
\end{align}
where $\mathbf{v}_{T}^{F(k)}$ and $\mathbf{v}_{S}^{(k)}$ denote the $k$-th individual eigenvector 1D arrays extracted from $\mathbf{V}_T^F$ and $\mathbf{V}_S$, respectively, with $D_F = 6$ corresponding to the fixed soft-biometric categories (gender, race, age, skin tone, body build, and facial structure).
This objective operates via two complementary mechanisms.
The first term explicitly aligns the geometric axes of the latent spaces by maximizing the squared cosine similarity between the principal components of the teacher and the student.
The second term aligns the actual data distributions along these axes by minimizing the Mean Squared Error between the projected semantic and visual coordinates.
Together, these mechanisms precisely transfer core invariant biometric intelligence while filtering out the MLLM's contextual background noise.

\begin{figure}
    \centering
    \includegraphics[width=\linewidth]{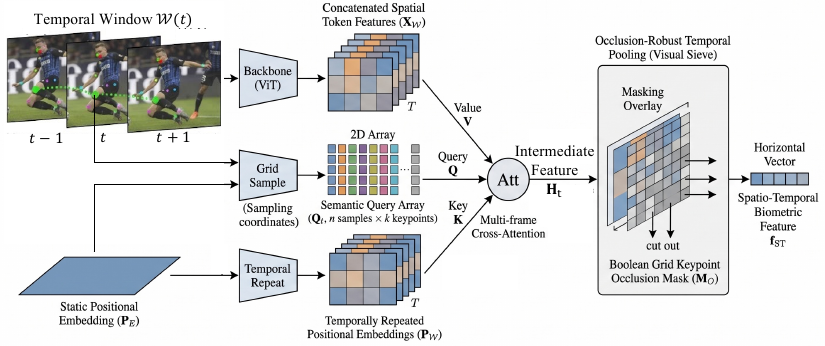}
    \vspace{-1.75em}
    \caption{\small Illustration of K-SAH.
    It extends static semantic attention to model local temporal dynamics over a temporal window \(\mathcal{W}\).
    Semantic queries \(\mathbf{Q}_t\) for the anchor frame $t$ are derived via grid sampling from static embeddings, preserving spatial priors.
    Cross-frame attention between the queries \(\mathbf{Q}_t\) and dense spatiotemporal key \(\mathbf{P}_W\) and value \(\mathbf{X}_W\) volumes allows precise tracking of geometric trajectories.
    Occlusion-Robust Temporal Pooling dynamically aggregates intermediate features \(\mathbf{H}_t\) by masking occluded patches using an explicit mask \(\mathbf{M}_O\), producing the final flattened spatiotemporal feature \(\mathbf{f}_{ST}\).
    \vspace{-3mm}}
    \label{fig:ksah}
\end{figure}

\paragraph{Transient Noise Disentanglement (TND).}
Dynamic soft biometrics must be explicitly decoupled from the identity representation to prevent reliance on transient attributes such as identical outfits.
We calculate the empirical covariance matrix for $\mathbf{X}^{D}$ and utilize SVD to extract its top $D_D$ principal components, forming the matrix $\mathbf{V}_T^D \in \mathbb{R}^{C \times D_D}$, which defines the geometric subspace of transient noise ($D_D = 5$ corresponding to hair, facial hair, accessories, upper clothing, and lower clothing).
We mathematically force the student's unified projected feature, $\mathbf{Y}$, to be orthogonal to this dynamic subspace via a direct projection penalty:
\begin{align}
    \mathcal{L}_{TND} = \left\| \mathbf{Y} \mathbf{V}_{T}^{D} \right\|_F^2.
\end{align}

\paragraph{Relational Topology Preservation.}
To complement ITA, we preserve the relational topology of instances within a training batch by comparing the Gram matrices of the teacher and student features:
\begin{align}
    \mathcal{L}_{FR} = \frac{1}{N^2} \| \mathbf{X}^{F} (\mathbf{X}^{F})^\top - \mathbf{Y} \mathbf{Y}^\top \|_F^2.
\end{align}
This ensures that individuals sharing similar fixed soft biometrics in the language space have proportionally minimized distances in the metric space.

\subsection{Kinematic Semantic Attention Head (K-SAH)}

The primary challenge in extending human recognition to the temporal domain is the severe disparity in data availability.
The scale, diversity, and annotation quality of static image datasets (\eg, WebBody4M~\cite{kim2025sapiensid}) vastly exceed those of video datasets.
Consequently, training parameter-heavy spatiotemporal models from scratch on limited video data inevitably leads to severe overfitting and the destruction of rich, pre-trained spatial priors.

Rather than discarding these spatial priors or building a separate video architecture, we lift the pre-trained image foundation model into the temporal domain by introducing the Kinematic Semantic Attention Head (K-SAH).
This builds on SapiensID's spatial representations which inherently encode precise geometric locations via the RP positional embeddings.
Because the spatial tokens explicitly retain the coordinates of keypoints, we can natively track these trajectories over time, thereby capturing \textit{kinematic continuity} simply by expanding the attention window.

Crucially, this design guarantees complete checkpoint compatibility with the existing foundational model.
The pre-trained weights of the static SAH are directly transferable to K-SAH, allowing the network to leverage temporal information without architectural disruption.
In fact, the original SAH is mathematically a special case of K-SAH when the input sequence has a single frame ($T=1$).

\paragraph{Temporal Windowing and Spatio-Temporal Embeddings.}
Given a video tracklet yielding spatial feature tensor $\mathbf{X} \in \mathbb{R}^{B \times T \times L \times C}$ (where $B$ is the batch size), we explicitly model the local temporal dynamics without flattening the entire video sequence, which would cause quadratic attention bottlenecks.
For each frame $t$, we define a localized temporal window $\mathcal{W}(t)$.

We simply repeat the standard pre-trained 2D positional embedding matrix $\mathbf{P}$ across the temporal dimension.
This approach uses the physical displacement of the subject to provide kinematic context and preserves the spatial knowledge from WebBody4M image pre-training.
Let $\mathbf{P}_{\mathcal{W}} \in \mathbb{R}^{|\mathcal{W}| L \times C}$ denote this temporally repeated 2D positional embedding for the given window.

\paragraph{Multi-Frame Semantic Attention.}
Within K-SAH, the semantic query $\mathbf{Q}_t$ is generated for the anchor frame \(t\) via grid sampling of \(\mathbf{P}\) using its frame-level keypoint predictions \(\mathbf{C}_t\), similar to Eq.~\ref{eq:image_grid_sample}:
\begin{align}
    \mathbf{Q}_t = \text{GridSample}\left( \mathbf{P}, \mathbf{C}_t \right) + \mathbf{O}.
\end{align}
However, rather than attending solely to the spatial tokens of frame $t$ (as in Eq.~\ref{eq:image_semantic_attention}), the query attends to the concatenated spatial features $\mathbf{X}_{\mathcal{W}} \in \mathbb{R}^{|\mathcal{W}| L \times C}$ and their corresponding repeated 2D positional embeddings $\mathbf{P}_{\mathcal{W}}$ within the temporal window.
The multi-frame cross-attention yields an intermediate temporal feature matrix for all keypoints at frame $t$:
\begin{align}
    \mathbf{H}_{t} = \text{Attention}\left(\mathbf{Q}_{t}, \mathbf{P}_{\mathcal{W}}, \mathbf{X}_{\mathcal{W}}\right).
\end{align}
Unlike global spatio-temporal attention methods that scale quadratically $\mathcal{O}(T^2 L^2 C)$, K-SAH restricts cross-attention strictly along the $K$ keypoint trajectories, reducing complexity to linear scaling $\mathcal{O}(T \cdot K \cdot L \cdot C)$ with respect to window size $T$.
Because the feature space is completely compatible with the pre-trained static weights, this formulation replaces isolated spatial pooling without requiring video retraining.
It allows the attention mechanism to smoothly track specific semantic patches (\eg, a swinging arm or stepping foot) across neighboring frames, capturing continuous kinematics.

\paragraph{Occlusion-Robust Temporal Pooling.}
A critical challenge in unconstrained real-world video is occlusion, where keypoints frequently disappear and reappear throughout a tracklet.
K-SAH elegantly handles this via dynamically masked temporal pooling prior to final feature flattening.

Let $\mathbf{M}_O \in \{0, 1\}^{B \times T \times K}$ be a boolean mask indicating whether a keypoint index $k \in \{1, \dots, K\}$ is missing or severely occluded in frame $t$.
During training and inference, $\mathbf{M}_O[t, k] = 1$ is generated dynamically using OpenPose keypoint confidence scores when the confidence drops below a threshold ($<0.3$).
Furthermore, let $\mathbf{h}_{t, k}$ denote the specific 1D feature vector for keypoint $k$ extracted from the matrix $\mathbf{H}_t$.
We pool the intermediate multi-frame features $\mathbf{h}_{t, k}$ across the temporal dimension, explicitly disregarding occluded patches:
\begin{align}
    \bar{\mathbf{h}_{k}} = \frac{\sum_{t=1}^T \mathbf{h}_{t, k} \cdot (1 - \mathbf{M}_O[t, k])}{\max \left(1, \sum_{t=1}^T (1 - \mathbf{M}_O[t, k]) \right)}.
\end{align}
This ensures that heavily occluded frames do not corrupt the aggregated representation.
The temporally pooled features are then flattened to produce the final spatiotemporal feature $\mathbf{f}_{ST}$.

%% file: secs/4_experiments.tex
\section{Experiments}

\paragraph{Experiment Setup.}
Unless otherwise noted, our experimental setup, including the ViT backbone, RetinaPatch extraction, and training hyperparameters, strictly follows SapiensID~\cite{kim2025sapiensid}.
For the semantic distillation modules, we utilize \verb|Qwen2.5-VL-7B-Instruct|~\cite{bai2025qwen2} to extract the semantic subspaces offline, retaining the top principal components for both ITA ($D_F=6$) and TND ($D_D=5$).
For video-based tasks, K-SAH operates on randomly sampled temporal windows of $T=5$ frames.
We train a single model strictly on the WebBody4M dataset and evaluate it directly in a zero-shot manner on all testing benchmarks without dataset-specific fine-tuning.
For video-based evaluation on CCPG, we evaluate on RGB video sequences, extracting kinematic trajectories via K-SAH.
Because baseline models are bound to their input modalities (gait models strictly require video data like CCGR, while appearance models are trained on static corpora), all models are evaluated zero-shot on the target test sets for a fair comparison of out-of-distribution generalization (see Table 6 in the supplementary material for baseline pretraining metadata).
We use the same checkpoint for each baseline model.


\begin{table*}[t]
\centering
\caption{\small Generalization comparison with SoTA ReID and gait recognition models.
The evaluation is split into video-based person reID (CCVID) and gait recognition (CCPG).
For CCPG, we report the rank-1 accuracy under four evaluation protocols.
SapiensID 2.0 leverages semantic and kinematic priors to achieve SoTA performance.}
\label{tab:gait}
\vspace{1.75mm}

\hfill
\begin{minipage}[t]{0.49\linewidth}
\centering
\footnotesize

\begin{NiceTabular}{lcccc}
\toprule
\Block{2-1}{Method} & \Block{1-2}{CCVID (General)} & & \Block{1-2}{CCVID (CC)} & \\
\cmidrule(lr){2-3} \cmidrule(lr){4-5}
 & top1 & mAP & top1 & mAP \\
\midrule
CAL~\cite{gu2022clothes} & 74.83 & 29.43 & 73.89 & 26.65 \\
CLIP3DReID~\cite{liu2023learning} & 77.28 & 30.01 & 76.28 & 26.69 \\
SOLIDER~\cite{chen2023beyond} & 40.27 & 36.56 & 39.61 & 35.48 \\
HAP~\cite{yuan2023hap} & 54.74 & 45.14 & 52.37 & 41.33 \\ \midrule
BiggerGait~\cite{ye2025biggergait} & 85.13 & \underline{83.38} & 84.53 & \textbf{82.59} \\
\midrule
SapiensID~\cite{kim2025sapiensid} & \underline{92.57} & 77.82 & \textbf{88.72} & 72.22 \\
Ours & \textbf{95.70} & \textbf{84.99} & \underline{88.62} & \underline{78.04} \\
\bottomrule
\end{NiceTabular}
\end{minipage}\hfill
\begin{minipage}[t]{0.49\linewidth}
\centering
\footnotesize

\begin{NiceTabular}{lcccc}
\toprule
\Block{2-1}{Method} & \Block{1-4}{CCPG~\cite{li2023depth}} & & & \\
\cmidrule(lr){2-5}
 & CL & UP & DN & BG \\
 \midrule
CLIP3DReID~\cite{liu2023learning} & 15.7 & 25.6 & 24.5 & 60.3 \\\midrule
GaitBase~\cite{fan2023opengait} & 20.8 & 31.3 & 38.3 & 70.2 \\
DeepGaitV2~\cite{fan2023exploring} & 23.0 & 37.1 & 36.5 & 69.9 \\
BigGait~\cite{ye2024biggait} & 22.9 & 42.2 & 25.0 & 80.5 \\
BiggerGait~\cite{ye2025biggergait} & 23.2 & 44.5 & 29.5 & 86.7 \\
\midrule
SapiensID~\cite{kim2025sapiensid} & 33.9 & 45.0 & 75.2 & 90.2 \\
Ours & \textbf{60.8} & \textbf{60.8} & \textbf{86.4} & \textbf{91.6} \\
\bottomrule
\end{NiceTabular}
\end{minipage}\hfill
\end{table*}

\paragraph{Video-Based Person ReID and Gait Recognition.}
Table~\ref{tab:gait} evaluates SapiensID 2.0 on video-based person reID (CCVID~\cite{gu2022clothes}) and gait recognition (CCPG~\cite{li2023depth}).
The results clearly illustrate the divergence between models that overfit to domain-specific cues and those that learn invariant representations.
On CCVID, while video-specialized models like BiggerGait~\cite{ye2025biggergait} achieve strong overall accuracy, SapiensID 2.0 achieves top performance ($95.70\%$ General Rank-1).
More importantly, it demonstrates robustness in the challenging clothing-change (CC) protocol, boosting SapiensID's mAP by $+5.82\%$ (from $72.22\%$ to $78.04\%$).
This suggests that explicitly penalizing dynamic visual traits (via TND) encourages the model to synthesize a reliable temporal identity rather than simply tracking an outfit across frames.

The performance on CCPG further exposes the generalization bottlenecks of specialized architectures.
Models trained on ReID data (\eg, CLIP3DReID~\cite{liu2023learning}) inherently struggle to capture identity information from gait data.
Conversely, the poor cross-dataset generalization of gait models indicates they likely overfit to their training domains (\eg, BiggerGait reaches only $23.2\%$ Rank-1 on CCPG CL).
By leveraging K-SAH to explicitly track the temporal displacement of spatial patches through keypoints, SapiensID 2.0 extracts robust kinematic signatures, substantially improving performance on CCPG CL ($33.9\%$ to $60.8\%$) without requiring massive video retraining.

\begin{table}[]
\centering
\caption{\small Generalization comparison with SoTA ReID models across short-term and long-term settings.
``Long-term'' refers to clothing change (CC) protocols of LTCC and PRCC datasets, while ``short-term'' uses the same clothing (SC) protocol.
For other datasets, the data capture characteristics define short or long-term conditions.
While texture-biased models excel marginally on short-term benchmarks, SapiensID 2.0 achieves much superior robustness and generalization in long-term scenarios.}
\label{tab:wholebody}
\footnotesize
\begin{NiceTabular}{lccccccccc}
\toprule
\Block{2-1}{Method} & \Block{2-1}{Avg} & \Block{1-2}{LTCC (General)~\cite{shu2021large}} & & \Block{1-2}{PRCC (SC)~\cite{yang2019person}} & & \Block{1-2}{Market1501~\cite{zheng2015scalable}} & & \Block{1-2}{MSMT17~\cite{wei2018person}} & \\
\cmidrule(lr){3-4} \cmidrule(lr){5-6} \cmidrule(lr){7-8} \cmidrule(lr){9-10}
 & & top1 & mAP & top1 & mAP & top1 & mAP & top1 & mAP \\
\midrule
CAL~\cite{gu2022clothes} & 49.09 & 72.41& 38.12& 99.54& 99.01& 43.65& 21.03& 14.48& 4.44 \\
CLIP3DReID~\cite{liu2024distilling} & 50.20 & 75.66& \textbf{45.15}& 99.43& 96.43& 41.66& 20.33& 17.45& 5.50 \\
SOLIDER~\cite{chen2023beyond} & \textbf{71.45} & 73.83& 36.28& 99.51& \textbf{99.53} & \textbf{97.03}& \textbf{94.04}& 48.64& 22.77 \\
HAP~\cite{yuan2023hap} & 70.78 & 73.02& 35.97& 99.30& 98.45& 96.23& 92.20& 48.01& 23.02 \\
SapiensID~\cite{kim2025sapiensid} & 70.01 & 72.01& 34.56& \textbf{100.00}& 98.79& 88.18& 68.26& 67.25& \textbf{31.02} \\
\midrule
SapiensID 2.0 (Ours) & 71.31 & \textbf{77.08}& 42.39& \textbf{100.00}& 99.01& 86.94& 67.15& \textbf{67.62} & 30.32 \\
\bottomrule
\end{NiceTabular}

\vspace{1mm}
\textbf{(a) Short-Term ReID}
\vspace{2mm}

\begin{NiceTabular}{lccccccccc}
\toprule
\Block{2-1}{Method} & \Block{2-1}{Avg} & \Block{1-2}{LTCC (CC)~\cite{shu2021large}} & & \Block{1-2}{PRCC (CC)~\cite{yang2019person}} & & \Block{1-2}{CCDA~\cite{liu2023learning}} & & \Block{1-2}{Celeb-ReID~\cite{huang2019celebrities}} & \\
\cmidrule(lr){3-4} \cmidrule(lr){5-6} \cmidrule(lr){7-8} \cmidrule(lr){9-10}
 & & top1 & mAP & top1 & mAP & top1 & mAP & top1 & mAP \\
\midrule
CAL~\cite{gu2022clothes} & 24.26 & 33.16& 16.27& 45.39& 45.42& 3.74& 9.14& 37.11& 3.81 \\
CLIP3DReID~\cite{liu2024distilling} & 24.93 & 41.84& \textbf{22.58}& 40.81& 38.38& 4.31& 10.18& 37.31& 4.02 \\
SOLIDER~\cite{chen2023beyond} & 21.66 & 25.00& 12.18& 26.87& 32.12& 8.62& 16.48& 46.37& 5.66 \\
HAP~\cite{yuan2023hap} & 22.66 & 24.74& 11.71& 33.90& 37.00& 8.30& 16.02& 44.38& 5.20 \\
SapiensID~\cite{kim2025sapiensid} & 62.77 & 42.35& 17.79& 78.75& 72.60& 61.84& 69.08& \textbf{92.80}& 66.92 \\
\midrule
SapiensID 2.0 (Ours) & \textbf{66.82} & \textbf{48.26}& 22.28& \textbf{85.19}& \textbf{79.32}& \textbf{64.96}& \textbf{71.06}& 91.64& \textbf{71.87} \\
\bottomrule
\end{NiceTabular}

\vspace{1mm}
\textbf{(b) Long-Term ReID}
\end{table}

\paragraph{Image-Based Person ReID.}
We evaluate SapiensID 2.0 on image-based person reID benchmarks where the improvements come from the semantic alignment with MLLM.
The distinction between short-term (same clothing) and long-term (clothing change) protocols reveals a critical flaw in previous methods: short-term datasets inherently reward models that overfit to transient appearance cues.
As shown in Tab.~\ref{tab:wholebody}a, models like SOLIDER perform exceptionally well on short-term tasks (averaging $71.45\%$ accuracy).
We infer that they achieve this by exploiting texture and clothing.

This inference is supported by their performance drop in unconstrained, real-world conditions where subjects change outfits, causing SOLIDER's average performance to collapse to $21.66\%$ in long-term scenarios (Tab.~\ref{tab:wholebody}b).
In contrast, SapiensID 2.0 deliberately penalizes reliance on these transient cues via the TND objective.
While this occasionally causes our model to marginally underperform SOLIDER on short-term benchmarks (averaging $71.31\%$), it fundamentally anchors the latent space to invariant anatomical traits (via ITA).
Consequently, SapiensID 2.0 achieves massive, consistent robustness across long-term protocols, elevating the baseline average from $62.77\%$ to $66.82\%$, demonstrating robustness to appearance variations.



\begin{table}[t!]
    \centering
    \begin{minipage}{0.44\linewidth}
        \centering\footnotesize
        \caption{\small 1:1 Face verification accuracy across face recognition benchmarks.
        These results confirm that SapiensID 2.0 fully preserves SoTA facial recognition capabilities.}
        \label{tab:face}
        \resizebox{\linewidth}{!}{
        \begin{NiceTabular}{lccc}
            \toprule
            Method & \Block{1-1}{AdaFace \\ \cite{kim2022adaface}} & \Block{1-1}{SapiensID \\ \cite{kim2025sapiensid}} & \Block{1-1}{SapiensID \\ 2.0 (Ours)} \\
            \midrule
            LFW~\cite{huang2008labeled} & 99.82 & 99.82 & \textbf{99.85} \\
            CPLFW~\cite{zheng2018cross} & 95.12 & 94.85 & \textbf{95.28} \\
            CFPFP~\cite{sengupta2016frontal} & \textbf{99.19} & 98.74 & 99.14 \\
            CALFW~\cite{zheng2017cross} & \textbf{96.07} & 95.78 & 95.93 \\
            AgeDB~\cite{moschoglou2017agedb} & 97.97 & 97.33 & \textbf{98.07} \\ \midrule
            Avg & 97.63 & 97.30 & \textbf{97.65} \\
            \bottomrule
        \end{NiceTabular}
        }
    \end{minipage}
    \hfill
    \begin{minipage}{0.54\linewidth}
        \centering\footnotesize
        \caption{\small Component-wise ablation of SapiensID 2.0 on the CCVID dataset.
        ``Gen'' denotes the general evaluation protocol, ``CC'' denotes the clothing-change protocol.
        The results confirm that combining these components yields highly synergistic performance gains.}
        \label{tab:ablation_components}
        \begin{NiceTabular}{ccccccc}
            \toprule
            \Block{2-1}{ITA} & \Block{2-1}{TND} & \Block{2-1}{K-SAH} & \Block{1-2}{CCVID (Gen)} & & \Block{1-2}{CCVID (CC)} & \\ \cmidrule(lr){4-5} \cmidrule(lr){6-7}
            & & & top1 & mAP & top1 & mAP \\
            \midrule
            - & - & - & 92.57 & 77.82 & \textbf{88.72} & 72.22 \\
            \checkmark & - & - & 92.70 & 80.70 & 87.97 & 73.58 \\
            \checkmark & \checkmark & - & 93.82 & 81.26 & 88.39 & 75.66 \\
            \checkmark & \checkmark & \checkmark & \textbf{95.70} & \textbf{84.99} & {88.62} & \textbf{78.04} \\
            \bottomrule
        \end{NiceTabular}
    \end{minipage}
\end{table}


\paragraph{Face Recognition.}
To verify that optimizing for body-centric soft biometrics and temporal kinematics does not induce catastrophic forgetting of fine-grained facial features, we evaluate SapiensID 2.0 across five standard face recognition benchmarks.
As shown in Tab.~\ref{tab:face}, SapiensID 2.0 fully preserves, and slightly exceeds, SoTA face verification accuracy (e.g., $99.85\%$ on LFW vs.\ AdaFace-ViT's $99.82\%$).
We attribute this to ITA.
We hypothesize that because the invariant soft biometrics extracted from the MLLM (\eg gender) naturally correlate with structural facial features, the body-centric semantic alignment successfully avoids inducing catastrophic forgetting of fine-grained facial priors.

\paragraph{Ablation Studies.}
To systematically isolate the contributions of our proposed semantic distillation and temporal tracking modules, we conduct ablation studies on CCVID~\cite{gu2022clothes}.
It is suited for this analysis as it provides video tracklets and separate protocols for general (short-term) and clothing-change (long-term) scenarios.
Tab.~\ref{tab:ablation_components} reports the progressive integration of our modules over the semantically agnostic, static baseline.
Introducing ITA provides an immediate boost to both general and clothing-change mAP (\eg, General mAP jumps from $77.82\%$ to $80.70\%$), suggesting that anchoring the geometric space to stable semantic priors (\eg, body proportions, gender) enhances overall discriminability.
The addition of TND yields a particularly pronounced improvement on the CCVID (CC) split, increasing CC mAP by a further $+2.08\%$.
By mathematically enforcing orthogonality against dynamic noise, TND actively prevents the model from overfitting to identical outfits, effectively mitigating the baseline's ``semantic blindness.''
Introducing K-SAH explicitly models temporal dynamics.
As shown in the last row, swapping the original SAH for K-SAH drives the final push to SoTA metrics, improving General mAP to $84.99\%$ and CC mAP to $78.04\%$.
Furthermore, evaluating the temporal window size $T \in \{1, 3, 5, 7\}$ in K-SAH on CCVID (see supplementary material) shows steady gains over the static baseline ($T=1$, $72.22\%$ CC mAP), with performance saturating at $T=5$ ($78.04\%$ CC mAP).

\paragraph{Efficiency.}
Because MLLM feature extraction is performed offline, a training iteration increases marginally from 1.77 seconds to 2.06 seconds compared to the baseline.
During inference, the only added computational cost is the temporal attention in the K-SAH module.
Total evaluation time increases modestly from 28 minutes to 32 minutes.
This demonstrates that our approach achieves SoTA performance without compromising practical deployment speeds.

\begin{figure}
    \centering
    \includegraphics[width=0.95\linewidth]{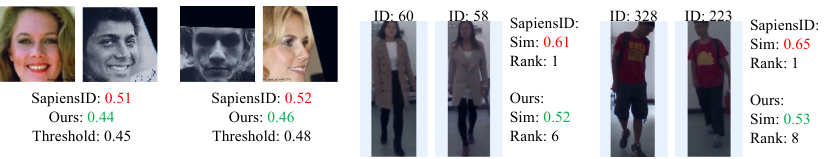}
    \vspace{-0.75em}
    \caption{\small Left: Compared to SapiensID, SapiensID 2.0 reduces the similarity between faces of different genders to below the verification threshold.
    Right: In retrieval, SapiensID returns another subject wearing similar clothing at rank-1.
    SapiensID 2.0 reduces the similarity between them to avoid making the same mistake.
    \vspace{-3mm}}
    \label{fig:visualization}
\end{figure}

\paragraph{Visualizations \& Discussion.}
To qualitatively validate our semantic distillation, Fig.~\ref{fig:visualization} illustrates cases where the baseline SapiensID fails but SapiensID 2.0 succeeds.
The baseline model suffers from ``semantic blindness,'' erroneously assigning high similarity to different individuals who share similar clothing (dynamic soft biometrics) or failing to distinguish clear differences in gender (fixed soft biometrics).
This visually confirms that SapiensID 2.0 effectively cures this blindness, successfully converting these false positives into true negatives, which is better aligned with human perception.
While learning clothing-invariant identity representations substantially improves generalization in realistic long-term scenarios, it incurs a slight, expected trade-off on short-term benchmarks (\eg, Market-1501) where clothing is a valid retrieval shortcut.
Additionally, potential attribute hallucinations from MLLMs in complex scenes are naturally mitigated in WebBody4M because the dataset is constructed via single-subject name queries, minimizing multi-person crop contamination.

%% file: secs/5_conclusion.tex
\section{Conclusion}

We present SapiensID 2.0, a unified foundation model that grounds human recognition in explicit semantic awareness and kinematic continuity.
We introduce a semantic distillation framework utilizing Bipartite Subspace Alignment.
This strategy leverages ITA and TND to anchor identity to stable anatomical traits while explicitly decoupling transient artifacts.
Furthermore, our K-SAH extends spatial attention to track structural patches over time, capturing continuous motion signatures without the prohibitive overhead of video-level retraining.
Extensive experiments demonstrate that SapiensID 2.0 achieves SOTA performance across image- and video-based person re-identification and gait recognition, while preserving robust face recognition capabilities.

\paragraph{Acknowledgements.}
This research is based upon work supported in part by the Office of the Director of National Intelligence (ODNI), Intelligence Advanced Research Projects Activity (IARPA), via 2022-21102100004. The views and conclusions contained herein are those of the authors and should not be interpreted as necessarily representing the official policies, either expressed or implied, of ODNI, IARPA, or the U.S. Government. The U.S. Government is authorized to reproduce and distribute reprints for governmental purposes notwithstanding any copyright annotation therein.

%% file: secs/X_supp.tex
\section{Additional Experiments}

\subsection{Additional Implementation Details}

\paragraph{Architecture and K-SAH Specifications}
SapiensID 2.0 is built upon a standard Vision Transformer (ViT-B/16) backbone, initialized with the pre-trained weights from the original SapiensID baseline to preserve foundational spatial representations. The RP module explicitly isolates and normalizes semantic regions (e.g., face, full body, upper torso) before patch extraction, preserving absolute geometric coordinates via region-specific positional embeddings. For our K-SAH, the temporal processing window size is set to $T=5$ frames during training to balance memory constraints and temporal representation learning. However, because the temporal pooling mechanism relies on attention masking and late sequence aggregation, this window can be flexibly scaled up during inference without further architectural modifications.

\paragraph{Semantic Subspace Distillation Setup}
The semantic feature spaces are generated entirely offline. We utilize the frozen \verb|Qwen2.5-VL-7B-Instruct|~\cite{bai2025qwen2} model to process the massive WebBody4M dataset. We employ the following structured prompt to ensure attribute consistency across instances: 
\begin{verbatim}
"Analyze the individual in the provided image or video tracklet. 
Generate a strict JSON object with:\n"
"group_a_fixed: race_ethnicity_presentation, gender_presentation, 
age_demographic_category, "
"skin_tone, body_shape, facial_structure.\n"
"group_b_dynamic: hair_color_length_style, facial_hair_presence, 
visible_accessories, "
"upper_clothing_color_type, lower_clothing_color_type.\n"
"If an attribute is obscured or ambiguous, return null for that key. "
"Do not infer uncertain information. Return JSON only."
\end{verbatim}

The resulting text strings are passed through the \verb|Qwen2.5-VL-7B-Instruct|~\cite{bai2025qwen2} text encoder, producing high-dimensional textual embeddings. To obtain the requisite stable semantic topologies, we compute the covariance matrix over the entire WebBody training corpus and apply SVD independently to both the fixed and dynamic semantic sets. For ITA, we retain the top $D_F=6$ principal components from the fixed biometrics subspace ($\mathbf{V}_T^F$). For TND, we extract the top $D_D=5$ principal components from the dynamic biometrics subspace ($\mathbf{V}_T^D$). To align the visual embedding space with these lower-dimensional semantic bases, we attach a lightweight, two-layer Multi-Layer Perceptron (MLP) mapping the student feature dimension ($D \rightarrow 512 \rightarrow D$). This MLP serves exclusively to calculate the distillation gradients during training and is dropped prior to inference.

\paragraph{Training Hyperparameters}
The unified SapiensID 2.0 framework is optimized using the AdamW optimizer. We employ a base learning rate of $1 \times 10^{-4}$, accompanied by a cosine annealing learning rate schedule and a linear warmup over the first two epochs. Weight decay is set to $0.05$. To effectively support the Relational Topology Preservation loss ($\mathcal{L}_{FR}$), we utilize a large total batch size of $128$, which is distributed across 7 NVIDIA A100 GPUs. The model is trained for a total of 30 epochs. For the loss landscape, we use AdaFace~\cite{kim2022adaface}, weighting our combined semantic distillation loss terms equally with a projection weight $0.5$ to ensure that the distillation process guides the latent space without overriding the fundamental metric learning constraints.

\subsection{Datasets}

\begin{table}[]
    \caption{Summary of testing datasets used in our experiments.}
    \label{tab:datasets}
    \centering\small
   \begin{NiceTabular}{lcccccc}
        \toprule
        Dataset & Year & Task & Data Type & Same-Cloth. & Cloth.-Change & Other Variations \\
        \midrule
        CCVID~\cite{gu2022clothes} & 2022 & ReID & Video & \checkmark & \checkmark & -- \\
        CCPG~\cite{li2023depth} & 2023 & ReID/Gait & Video &  & \checkmark & Gait \\
        LTCC~\cite{shu2021large} & 2021 & ReID & Image & \checkmark & \checkmark & -- \\
        PRCC~\cite{yang2019person} & 2019 & ReID & Image & \checkmark & \checkmark & -- \\
        Market1501 & 2015 & ReID & Image & \checkmark &  & -- \\
        MSMT17 & 2018 & ReID & Image & \checkmark &  & -- \\
        CCDA~\cite{liu2023learning} & 2023 & ReID & Image &  & \checkmark & -- \\
        CelevReID~\cite{huang2019celebrities} & 2019 & ReID & Image &  & \checkmark & -- \\
        WebBody~\cite{kim2025sapiensid} & 2025 & ReID & Image &  &  & Pose/Scale \\
        \midrule
        LFW & 2007 & Face & Image &  &  & -- \\
        CPLFW & 2018 & Face & Image &  &  & Pose \\
        CFP-FP & 2017 & Face & Image &  &  & Frontal \\
        CALFW & 2017 & Face & Image &  &  & Age \\
        AgeDB & 2017 & Face & Image &  &  & Age \\
        \bottomrule
    \end{NiceTabular}
\end{table}

In Tab.~\ref{tab:datasets}, we provide an overview of all the testing datasets that we use in our experiments. Note that we use one single model to test the zero-shot generalization capabilities of both SapiensID 2.0 and the baseline methods. 

\subsection{Additional Results}

\begin{table}[]
\centering
\caption{Generalization comparison with SoTA ReID models on two settings. 
"Long-term" refers to clothing change (CC) protocols of LTCC and PRCC datasets, while ``short-term'' the same clothing (SC) protocols. 
For other datasets, the data capture characteristics define short or long-term conditions. 
}
\label{tab:wholebody_full}
\footnotesize
\resizebox{\textwidth}{!}{
\begin{NiceTabular}{llcccccccc}
\toprule
\Block{2-1}{Method} & \Block{2-1}{Train Data} & \Block{1-2}{LTCC (General)} & & \Block{1-2}{PRCC (SC)~\cite{yang2019person}} & & \Block{1-2}{Market1501} & & \Block{1-2}{MSMT17~\cite{wei2018person}} & \\
\cmidrule(lr){3-4} \cmidrule(lr){5-6} \cmidrule(lr){7-8} \cmidrule(lr){9-10}
 & & top1 & mAP & top1 & mAP & top1 & mAP & top1 & mAP \\
\midrule
CAL~\cite{gu2022clothes} (R50) & LTCC & 74.04& 40.84& 99.51& 95.64& 35.60& 16.11& 15.92 & 5.06 \\
CAL~\cite{gu2022clothes} (R50) & PRCC & 20.69& 6.19& \textbf{100.00}& 99.76& 18.97& 6.47& 2.56& 0.69 \\
CAL~\cite{gu2022clothes} (R50) & LTCC+PRCC & 72.41& 38.12& 99.54& 99.01& 43.65& 21.03& 14.48& 4.44 \\
CLIP3DReID~\cite{liu2024distilling} (R50) & LTCC & 75.66& \textbf{45.15}& 99.43& 96.43& 41.66& 20.33& 17.45& 5.50 \\
CLIP3DReID~\cite{liu2024distilling} (R50) & PRCC & 21.30& 6.19& \textbf{100.00}& \textbf{99.84}& 20.93& 7.49& 3.28& 0.85 \\
SOLIDER~\cite{chen2023beyond} (Swin-Base) & LUP4M+Market1501 & 73.83& 36.28& 99.51& \textbf{99.53} & \textbf{97.03}& \textbf{94.04}& 48.64& 22.77 \\
SOLIDER~\cite{chen2023beyond} (Swin-Base) & LUP4M+MSMT17 & 74.44& 36.74& 99.30& 98.71& 89.85& 73.20& \textbf{91.12}& \textbf{78.01} \\
HAP~\cite{yuan2023hap} (ViT-Base) & LUP4M+LTCC & 65.11& 29.02& 95.53& 86.44& 51.63& 27.29& 20.89& 6.56 \\
HAP~\cite{yuan2023hap} (ViT-Base) & LUP4M+PRCC & 63.29& 29.36& 98.84& 98.38& 73.49& 50.11& 29.61& 10.99 \\
HAP~\cite{yuan2023hap} (ViT-Base) & LUP4M+Market1501 & 73.02& 35.97& 99.30& 98.45& 96.23& 92.20& 48.01& 23.02 \\
HAP~\cite{yuan2023hap} (ViT-Base) & LUP4M+MSMT17 & 67.95& 32.07& 99.15& 96.50& 80.37& 57.07& 89.13& 75.85 \\
HAP~\cite{yuan2023hap} (ViT-Base) & WebBody & 56.80& 25.88& 99.72& 98.26& 66.18& 42.41& 43.61& 21.42 \\
SapiensID~\cite{kim2025sapiensid} (ViT-Base) & WebBody & 72.01& 34.56& \textbf{100.00}& 98.79& 88.18& 68.26& 67.25& 31.02 \\
\midrule
SapiensID 2.0 (Ours, ViT-Base) & WebBody & \textbf{77.08}& 42.39& \textbf{100.00}& 99.01& 86.94& 67.15& 67.62& 30.32 \\
\bottomrule
\end{NiceTabular}
}

\vspace{1mm}
\textbf{(a) Short-Term ReID}
\vspace{2mm}

\resizebox{\textwidth}{!}{
\begin{NiceTabular}{llcccccccc}
\toprule
\Block{2-1}{Method} & \Block{2-1}{Train Data} & \Block{1-2}{LTCC (CC)~\cite{shu2021large}} & & \Block{1-2}{PRCC (CC)} & & \Block{1-2}{CCDA~\cite{liu2023learning}} & & \Block{1-2}{Celeb-ReID~\cite{huang2019celebrities}} & \\
\cmidrule(lr){3-4} \cmidrule(lr){5-6} \cmidrule(lr){7-8} \cmidrule(lr){9-10}
 & & top1 & mAP & top1 & mAP & top1 & mAP & top1 & mAP \\
\midrule
CAL~\cite{gu2022clothes} (R50) & LTCC & 38.01& 18.84& 37.00& 35.20& 3.91& 9.67& 37.42& 3.92 \\
CAL~\cite{gu2022clothes} (R50) & PRCC & 6.38& 3.14& 55.69& 55.64& 2.85& 8.61& 23.59& 2.20 \\
CAL~\cite{gu2022clothes} (R50) & LTCC+PRCC & 33.16& 16.27& 45.39& 45.42& 3.74& 9.14& 37.11& 3.81 \\
CLIP3DReID~\cite{liu2024distilling} (R50) & LTCC & 41.84& \textbf{22.58}& 40.81& 38.38& 4.31& 10.18& 37.31& 4.02 \\
CLIP3DReID~\cite{liu2024distilling} (R50) & PRCC & 6.63& 3.17& 62.40& 61.97& 3.17& 8.89& 23.82& 2.17 \\
SOLIDER~\cite{chen2023beyond} (Swin-Base) & LUP4M+Market1501 & 25.00& 12.18& 26.87& 32.12& 8.62& 16.48& 46.37& 5.66 \\
SOLIDER~\cite{chen2023beyond} (Swin-Base) & LUP4M+MSMT17 & 26.02& 11.33& 22.27& 25.36& 8.79& 15.54& 47.95& 6.14 \\
HAP~\cite{yuan2023hap} (ViT-Base) & LUP4M+LTCC & 25.00& 11.63& 26.14& 22.34& 4.56& 11.18& 30.28& 3.54 \\
HAP~\cite{yuan2023hap} (ViT-Base) & LUP4M+PRCC & 29.08& 12.52& 38.05& 41.94& 5.13& 13.40& 37.79& 4.48 \\
HAP~\cite{yuan2023hap} (ViT-Base) & LUP4M+Market1501 & 24.74& 11.71& 33.90& 37.00& 8.30& 16.02& 44.38& 5.20 \\
HAP~\cite{yuan2023hap} (ViT-Base) & LUP4M+MSMT17 & 23.47& 10.74& 23.82& 25.00& 6.27& 13.33& 46.37& 5.77 \\
HAP~\cite{yuan2023hap} (ViT-Base) & WebBody & 22.70& 9.96& 54.93& 49.38& 28.80& 41.49& 65.78& 18.93 \\
SapiensID~\cite{kim2025sapiensid} (ViT-Base) & WebBody & 42.35& 17.79& 78.75& 72.60& 61.84& 69.08& \textbf{92.80}& 66.92 \\
\midrule
SapiensID 2.0 (Ours, ViT-Base) & WebBody & \textbf{48.26}& 22.28& \textbf{85.19}& \textbf{79.32}& \textbf{64.96}& \textbf{71.06}& 91.64& \textbf{71.87} \\
\bottomrule
\end{NiceTabular}
}

\vspace{1mm}
\textbf{(b) Long-Term ReID}
\end{table}

We show the results of additional baseline models with more diverse architectures trained on more diverse training datasets in Tab.~\ref{tab:wholebody_full}. In the main paper, we report the results of the baseline model with the best overall generalization performance.

\subsection{Performance of Gait Recognition Models on CCVID}

To provide a comprehensive analysis of cross-domain capabilities, we evaluated recent large-scale gait foundation models, specifically BigGait~\cite{ye2024biggait} and BiggerGait~\cite{ye2025biggergait}, on the video-based person re-identification dataset (CCVID). We use models trained on CCGR~\cite{zou2024cross}, a large-scale gait recognition dataset commonly used for pretraining gait models with better generalization results than other gait datasets.

%
Surprisingly, gait recognition models demonstrate remarkably strong generalization to video-based ReID. BiggerGait achieves $85.13\%$ Rank-1 and $83.38\%$ mAP on the CCVID General split, vastly outperforming traditional ReID models like CAL ($74.83\%$ Rank-1) and CLIP3DReID ($77.28\%$ Rank-1). Most notably, in the clothing-change (CC) split, BiggerGait achieves a highly competitive mAP of $82.59\%$. This highlights the intrinsic advantage of gait architectures in appearance-altering scenarios: by relying entirely on kinematic silhouettes and discarding RGB texture, they are inherently immune to the clothing-change artifacts that catastrophically disrupt traditional ReID models like SOLIDER. 

Despite this resilience, dedicated gait models are ultimately constrained by their modality limit. Because they discard all visual semantics (such as facial features, fixed soft biometrics, and unoccluded visual markers), they fail to match the peak discriminative capabilities of foundation models for human recognition. SapiensID 2.0 significantly outperforms BiggerGait in Top-1 accuracy across both the General ($95.70\%$ vs. $85.13\%$) and CC ($88.62\%$ vs. $84.53\%$) splits. This demonstrates that while BiggerGait is robust against noise, the optimal strategy for human recognition is a unified approach. By integrating kinematic continuity (via K-SAH) with explicit visual soft biometrics (via Invariant Trait Alignment), SapiensID 2.0 captures the temporal robustness of gait models without sacrificing the rich, discriminative power of static semantic identity.

\subsection{Cross Pose-Scale ReID}

\begin{table}[]
\centering
\footnotesize
\caption{ReID Performance on variable pose and scale settings on the WebBody test set.}
\label{tab:web_body_test}
\begin{NiceTabular}{lcc}
\toprule
{Method} & top1 & mAP\\
\midrule
CAL~\cite{gu2022clothes} & 6.57 & 1.02 \\
SOLIDER~\cite{chen2023beyond} & 9.95 & 1.98 \\
HAP~\cite{yuan2023hap} & 8.22 & 1.52 \\
SapiensID~\cite{kim2025sapiensid} & 76.82 & 52.00 \\
\midrule
SapiensID 2.0 (Ours) & 76.94 & 52.16 \\
\bottomrule
\end{NiceTabular}
\end{table}

To ensure that semantic alignment and kinematic tracking do not degrade the foundational spatial representations of the baseline, we evaluate the model on the challenging WebBody~\cite{kim2025sapiensid} test set, which features extreme variations in pose and scale. As shown in Table~\ref{tab:web_body_test}, traditional ReID models completely collapse (\eg, SOLIDER yields only $9.95\%$ Rank-1 accuracy). The SapiensID baseline exhibits strong pose and scale invariance due to its massive pre-training on WebBody4M. Crucially, SapiensID 2.0 maintains and even slightly improves upon this robustness ($76.94\%$ Rank-1 and $52.16\%$ mAP). This confirms that our Bipartite Subspace Alignment successfully disentangles semantic features without compromising the network's underlying geometric spatial priors.




\subsection{Additional Ablation Studies and Benchmarks}

\paragraph{Temporal Window Size Ablation in K-SAH}
To further dissect the impact of kinematic continuity, Tab.~\ref{tab:ablation_window} illustrates the effect of the temporal window size ($T$) within K-SAH on the CCVID dataset. When $T=1$, the architecture effectively degenerates to the baseline static attention (SAH). As the temporal window expands ($T \ge 3$), the network successfully tracks keypoint-guided semantic patches across neighboring frames, capturing discriminative kinematic signatures that single-frame averaging completely discards. Performance saturates around $T=3\text{--}5$, demonstrating that local temporal windows sufficiently capture kinematic continuity without incurring unnecessary computation.

\begin{table}[h]
\centering
\small
\caption{Impact of temporal tracking window size ($T$) within the K-SAH module on the CCVID dataset.}
\begin{NiceTabular}{lcccc}
\toprule
\Block{2-1}{Window Size ($T$)} & \Block{1-2}{CCVID (General)} & & \Block{1-2}{CCVID (CC)} & \\
\cmidrule(lr){2-3} \cmidrule(lr){4-5}
& top1 & mAP & top1 & mAP \\
\midrule
1 (Static SAH) & 92.57 & 77.82 & 88.72 & 72.22 \\
3 & 95.70 & 84.99 & 88.62 & 78.04 \\
5 & 95.65 & 84.95 & 88.53 & 77.95 \\
7 & 95.73 & 84.90 & 88.54 & 77.87 \\
\bottomrule
\end{NiceTabular}
\label{tab:ablation_window}
\end{table}

\paragraph{Edge Latency and Inference Throughput Benchmark}
We benchmark SapiensID 2.0 against edge MLLMs on an NVIDIA H100 GPU (fp16, batch size 1). As shown in Tab.~\ref{tab:edge_latency}, edge MLLMs (MobileVLM V2~\cite{yin2024mobilevlm} and FastVLM~\cite{apple2024fastvlm}) suffer from low throughput due to autoregressive decoding of attribute tokens ($>5$s latency per frame). In contrast, SapiensID 2.0 completely discards the teacher MLLM after training, running inference strictly on the 63M parameter ViT-B backbone in a single forward pass with a latency of only 23.3ms (42.9 FPS), achieving a $225\times$ throughput speedup.

\begin{table}[h]
\centering
\small
\caption{Inference latency, throughput, and parameter comparison with edge MLLMs.}
\label{tab:edge_latency}
\begin{NiceTabular}{lcccc}
\toprule
Model & Parameters (Inference) & GFLOPs / Frame & Throughput (FPS) & Edge Latency \\
\midrule
MobileVLM V2 (1.7B) & 1.67B & 3221.2 & 0.190 & 5.27s \\
FastVLM (1.5B) & 1.68B & 1967.3 & 0.162 & 6.17s \\
\textbf{SapiensID 2.0 (Ours)} & \textbf{63M} (backbone) & \textbf{68.8} & \textbf{42.9} & \textbf{23.3ms} \\
\bottomrule
\end{NiceTabular}
\end{table}

\paragraph{Zero-Shot Out-of-Distribution Baseline Comparison}
To further evaluate out-of-distribution (OOD) clothes-changing generalization, Tab.~\ref{tab:ood_baselines} compares SapiensID 2.0 directly against specialized domain-specific zero-shot methods on the LTCC and PRCC clothing-change protocols. SapiensID 2.0 significantly outperforms recent zero-shot methods (e.g., $48.26\%$ vs. $11.7\%$ on LTCC, and $85.19\%$ vs. $81.0\%$ on PRCC), demonstrating the superior generalization of our decoupled semantic subspace alignment.

\begin{table}[h]
\centering
\small
\caption{Zero-shot out-of-distribution generalization comparison on LTCC (CC) and PRCC (CC).}
\label{tab:ood_baselines}
\begin{NiceTabular}{llcccc}
\toprule
\Block{2-1}{Method} & \Block{2-1}{Venue \& Year} & \Block{1-2}{LTCC (CC)} & & \Block{1-2}{PRCC (CC)} & \\
\cmidrule(lr){3-4} \cmidrule(lr){5-6}
& & Rank-1 (\%) & mAP (\%) & Rank-1 (\%) & mAP (\%) \\
\midrule
CAL~\cite{gu2022clothes} & CVPR 2022 & -- & -- & 37.5 & 35.4 \\
AIM~\cite{yang2023aim} & CVPR 2023 & -- & -- & 40.7 & 38.4 \\
GEFF~\cite{geff2024} & WACV 2024 & 11.7 & 4.8 & 81.0 & 50.5 \\
DLCR~\cite{dlcr2025} & WACV 2025 & -- & -- & 45.5 & 38.8 \\
\textbf{SapiensID 2.0 (Ours)} & \textbf{NeurIPS 2026} & \textbf{48.26} & \textbf{22.28} & \textbf{85.19} & \textbf{79.32} \\
\bottomrule
\end{NiceTabular}
\end{table}

\section{Limitations}

While SapiensID 2.0 significantly advances human recognition, it presents a few limitations. First, our semantic distillation framework intrinsically bounds the model's semantic awareness to the zero-shot capabilities of the teacher MLLM. Consequently, any inherent limitations or failure modes of the MLLM in recognizing nuanced soft biometrics will naturally propagate to the student embedding space. Second, extracting the semantic subspaces requires running the heavy MLLM on the entire training corpus offline, which introduces a substantial, albeit one-time, computational overhead during the dataset preparation phase. However, since the distillation projection and the teacher MLLM are entirely discarded during inference, the deployed visual foundation model maintains the high efficiency and low latency required for real-time edge applications without any loss in throughput.

\section{Potential Societal Impacts}

The deployment of highly accurate, modality-unified human recognition systems carries profound societal implications. On the positive side, robust identity matching across extreme poses, occlusions, and clothing changes can drastically improve public security, aid in rigorous forensic investigations, and facilitate the recovery of missing persons in highly unconstrained environments. Conversely, the increased efficacy of these systems raises valid concerns regarding privacy infringement and the potential for unwarranted mass surveillance. Furthermore, because our semantic distillation relies on an MLLM to anchor traits such as race and gender, there is a distinct risk of inheriting and amplifying systemic biases present in the teacher model's pre-training data, which could lead to discriminatory profiling. It is imperative that the deployment of such foundational biometric models is accompanied by strict ethical guidelines, rigorous bias auditing, and transparent regulatory oversight to prevent misuse and ensure equitable societal outcomes.